\documentclass[letterpaper, 10 pt, conference]{ieeeconf}  

\IEEEoverridecommandlockouts                              

\usepackage{graphicx}
\graphicspath{ {./images/} }
\newcommand{\rpm}{\raisebox{.2ex}{$\scriptstyle\pm$}}
\usepackage{graphicx}

\usepackage{cite}
\makeatletter
\let\NAT@parse\undefined
\makeatother
\usepackage{hyperref}

\title{
Identifying  Multiple Interaction Events from Tactile Data during Robot-Human Object Transfer  
}

\author{Mohammad-Javad Davari$^{1\dag}$, Michael Hegedus$^{1}$, Kamal Gupta$^{1*}$ and Mehran Mehrandezh$^{2}$
\thanks{*This work was supported by an NSERC Discovery Grant to Kamal Gupta.\hfill\break
$\dag$ In memoriam. Mohammad recently passed away.}
\thanks{$^{1}$School of Engineering Science,
        Simon Fraser University, Canada
        {\tt\small \{mdavarid,kamal,mehegdus\}@sfu.ca}}%
\thanks{$^{2}$Faculty of Engineering and Applied science,  
University of Regina, Canada
       {\tt\small mehran.mehrandezh@uregina.ca}}%
\thanks{20XX IEEE.  Personal use of this material is permitted.  Permission from IEEE must be obtained for all other uses, in any current or future media, including reprinting/republishing this material for advertising or promotional purposes, creating new collective works, for resale or redistribution to servers or lists, or reuse of any copyrighted component of this work in other works.}%
}

\begin{document}

\maketitle
\thispagestyle{empty}
\pagestyle{empty}

\begin{abstract}
 During a robot to human object handover task, several intended or unintended events may occur with the object - it may be  pulled, pushed, bumped or simply held - by the human receiver. We show that it is possible to differentiate between these events solely via tactile sensors. Training data from tactile sensors were recorded during interaction of human subjects with the object held by a 3-finger robotic hand. A Bag of Words approach was used to automatically extract effective features from the tactile data. A Support Vector Machine was used to distinguish between the four events with over 95 percent  average accuracy.
\end{abstract}


\section{INTRODUCTION}
  
Robots can aid humans in different ways. A common example that occurs in a variety of scenarios, such as in assisted care or more generally, in collaborative robotics, is  that of transferring an object from a robot to human, i.e., robot-human object handover task. A key aspect of the  handover task is the human receiver's interaction with the object, while the object is held in the robot's hand.  For instance,  during object handover, the receiver may pull the object as he/she grasps it, or, the receiver may accidentally bump the object. Such events may cause potential slippage of the object. Identifying and classifying these events are essential to the  proper and robust execution of the object handover task, since the robot must take appropriate action in response to such events. We have identified four such events, namely holding, bumping, pulling and  pushing that may occur during object handover, and in this paper,  we  focus on identifying these four events using tactile sensors only.    

Indeed humans are good at human-human handover task, and use multiple sensing modalities  including haptic, force, visual and audio \cite{strabala2013toward}. Some of these modalities have been used in robotics to perform handover tasks as well. For instance, in \cite{moon2014meet} authors used vision to adjust timing of the handover. In \cite{chan2012grip}, strain-gauge-based force sensors were used to model handover grip force. In this paper, we show that it is possible to identify the above mentioned four events solely based on data obtained via tactile sensors. We achieve this by using a machine learning based feature identification algorithm, Bag of Words (BoW) \cite{gui2014time}, on the tactile data stream, followed by a Support Vector Machine (SVM) classifier \cite{vapnik1998support}. 

The tactile data is essentially a time series, i.e., a sequence of pressure readings from a set of spatially distributed tactile elements (taxels). It has both temporal and spatial aspects that can be exploited for detecting features. Temporal information is intrinsically relevant for detecting the events since they will result in change in sensed pressure, hence, we focused on temporal features primarily, because utilizing spatial features on top will result in more parameters and thus needing more training data.

 To generate features, one can either hand-design the features or automatically generate them. We indeed first tested hand-designed features to test classification feasibility. This resulted in very poor results (more detail in section \ref{sec:3}), due to high variability in temporo-spatial tactile data encountered in the data collection process, even for the same subject. This motivated our decision to use automatic feature generation, the methods of choice for which are \cite{langkvist2014review}: Neural Networks (NN), Hidden Markov Model (HMM) and BoW. BoW uses the data itself as features but NN and HMM use the data to train their parameters. Thus they generally require more training data to generate effective features compared to BoW. It is onerous and  time consuming to collect a large amount of real data in our case, since our experiments require human subjects, and therefore, we favor the BoW approach to generate features. Furthermore, we chose SVM as a classifier because it requires data near the class boundaries (not the whole data set) \cite{bishop2016pattern}, thus it has a better chance of succeeding with less data.
  
 Initially, simply for feasibility purposes, the BoW+SVM approach was tested on data obtained from one subject only (30 training samples for two classes) with acceptable results. Following this initial experimental confirmation, we increased the number of participants to five and added various test objects for handover task such as a ball,  can and cuboid, in different grasp orientations to create variability via realistic scenarios. For example, through experiments, we observed that the  contact forces and contact locations on tactile pads vary widely  between a ball and a cuboid test object, which leads to a significantly more challenging scenario for the classifier. 
 
 Our paper presents results from five student volunteers interacting with a  3-finger mechanical Schunk Dexterous Hand (SDH) mounted on a six-DOF robotic arm.  Each finger has two sensor pads, a proximal and a distal, giving us a total of six tactile sensor pads. The sensor pads are commercially-available resistive  tactile sensors\cite{weiss}. Our approach obtained over 95\% accuracy with a 0.5 second window of tactile data, hence it is suitable  for real-time detection of these events. In summary, the key contributions of our work are: i) we show that it is possible to  classify if an object held in a robot's hand is being pushed, pulled, held, or bumped by a human receiver during a robot-human object transfer task, using only tactile data and standard machine learning techniques, ii) we present a single classifier for automatic detection of these multiple events - push, pull, bump or hold - using a BoW algorithm for automated feature generation on the temporal tactile data stream, followed by an SVM for classification, and iii) the detection is real time and hence can be used by the robot to ensure robust object transfer.
 
The organization of the paper is as follows. We describe related work in Section \ref{sec:1}, and detail nature of the data and the algorithms used in our work in Section \ref{sec:2}. The experimental setup is described in Section \ref{sec:3}, followed by the experimental procedure in Section \ref{sec:4} and the results in Section \ref{sec:5}. Finally, conclusion and future work are discussed in Section \ref{sec:6}.

\section{RELATED WORKS}\label{sec:1}

Most previous work in detecting events using tactile data is related to only a single event, either slip  detection \cite{francomano2013artificial,fernandez2016slip,holweg1996slip,alcazar2012estimating} or grasp stability detection \cite{bekiroglu2010learning,bekiroglu2011assessing,Haograsping2014}, primarily from a grasp stability perspective and does not directly concern itself with object handover. Our work, on the other hand, provides a single classifier for disparate multiple events that can occur during object handover task. The previous work, however, does give us information about the types of features used for tactile sensor data, as outlined below.

 Several spatial features have been used in the context of slip detection - from relatively simple ones like moments of 
 inertia \cite{bekiroglu2011assessing} to  more involved ones such as optical-flow-based techniques \cite{alcazar2012estimating}. \cite{bekiroglu2011assessing}  found that using temporal information  with a Hidden Markov Model (HMM) based classifier gives better results than a one shot classification based on spatial features such as  moments of inertia for grasp stability assessment. \cite{Haograsping2014} also used BoW for feature detection for grasp stability assessment, but only on spatial aspects, i.e., using only a single reading of sensors acquired at the end of grasping process. Optical flow based technique assumes that the object or point of contact with the tactile sensor is considerably smaller than the overall size of the tactile sensor, and furthermore they only work well with precise/high resolution tactile sensor. Frequency domain based techniques have also been used \cite{holweg1996slip,fishel2008robust,heyneman2016slip}, however they require sensors with a sampling rate much higher than that available on commercially available tactile sensors, such as the one we use. 
 
For completeness, we mention that tactile data has also been used to perform object recognition \cite{SchneideretalObjRecog2009, madry2014st}. The former uses BoW features but only on the spatial aspects. The latter does take into account temporal aspects of feature detection and  uses a neural network for feature generation. \cite{SohGaussianProObjRecog2012} used spatial statistical features such as moments along with an on-line learning algorithm that learns from the time series for object recognition. Both these works, however, do not deal with multiple event classification, as we do in our work.

\section{DATA TYPE AND ALGORITHM}\label{sec:2}

A tactile sensor typically is a grid of individual sensors, called taxels, that provide pressure readings at their respective location at a given frequency. The collected sensor data, therefore, consists of a stream of multiple taxel readings for a certain duration of time, represented as a multivariate time series data (MTSD) \cite{chakraborty2007feature}. One of the difficulties of working with MTSD is the issue of intra-class variability. This issue arises from the fact that the same event can occur at different speeds. To address these issues, methods like: BoW\cite{gui2014time}, recurrent neural networks \cite{husken2003recurrent}, dynamic time warping \cite{rakthanmanon2012searching} and HMM\cite{hermansky2000tandem} have been studied in the pertinent literature.  

Initially we hand designed features and tested them on a limited number of samples. These features were: maximum, mean, median and standard deviation of taxel values, frequency content via FFT, derivative mean, first moment, and optical flow. The best performance was obtained with the standard deviation feature at 60\% for pull vs grip classification which is clearly sub par.

For automatic feature generation, as mentioned earlier in the introduction, we chose BoW approach followed by an SVM classifier because it has a better chance of succeeding with less data. A simple description of BoW is given below.
 
Each experiment consists of multiple sequences (each sequence is of length T) of pressure readings from multiple tactile sensors. To compose a training data, only sequences with nonzero values are considered. The training data is given to the BoW algorithm as a series of tag-less sequences. Using a sliding window of length W, the BoW  divides these sequences into sub-sequences of length W. Each sub-sequence can be viewed as a point in the W-dimensional window-space. By performing K-means clustering within this space, K clusters (or features) are generated, each represented by its center. Since each training or test sequence has multiple sub-sequences, each sub-sequence represented by the closest center, will generate a histogram of centers for each sequence. This histogram is represented as a single point in the K-dimensional feature space then. For N training sequences, the output of the BoW algorithm includes N points in the feature space.

\begin{figure}[h]
\centering
\includegraphics[width=0.5\textwidth]{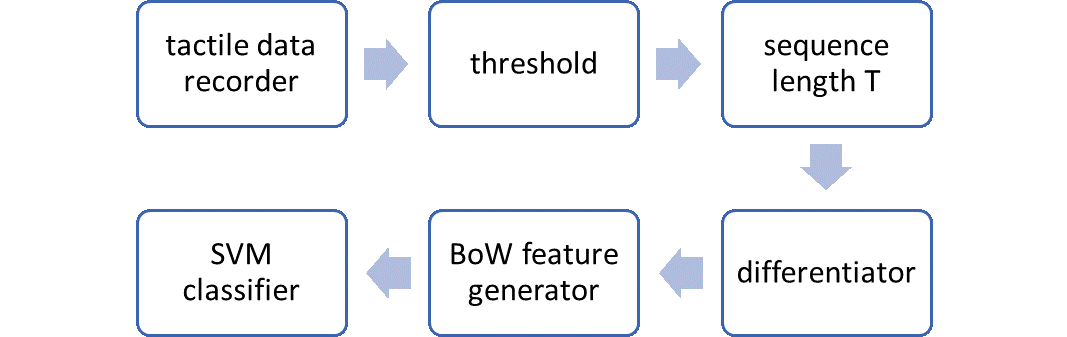}
\centering
\caption{Overall process of classification}
\centering
\label{algorithm_steps}
\end{figure}

In order to have a consistent data stream (among different experiments), we feed the derivative of the sensor data stream  to the BoW algorithm   as a pre-processing step, as shown in Fig ~\ref{algorithm_steps}. It eliminates constant biases in the data points, such as those due to gravity, due to a person's strength, or due to any variations among sensors, etc. 

To classify the sequences using the features provided, we used an SVM\cite{vapnik1998support} classifier. SVM classifies data by providing a hyperplane with the maximum margin from the closest data points from the opposing classes (i.e., support vectors). With the aid of an appropriate kernel, SVM is able to classify data with non-linear boundaries. The block diagram for the overall process of classification is shown in Figure \ref{algorithm_steps}.

\section{HARDWARE SETUP}\label{sec:3}

Our experimental setup consists of a three fingered Schunk Dexterous Hand (SDH), called robotic hand from now on. It has two resistive type sensor pads - DSA 9205 class tactile sensors \cite{weiss} -  per finger, one on the proximal and the other on the distal link.  Each pad has a 13x6 matrix of taxels for a total of 78 taxels per pad. Although the raw data at the taxel level are sampled at 230 HZ, the firmware compresses data by a factor of seven to provide a stable output consisting of data streams for all 6 pads, at 32 HZ. In our experiments, only precision grasps that involve distal part of the finger were utilized.  Hence only data from the 3 distal tactile sensors were collected in our studies. 

\begin{figure}[h]
\begin{tabular}{c c c}
 \includegraphics[width=.28\linewidth,height=60pt]{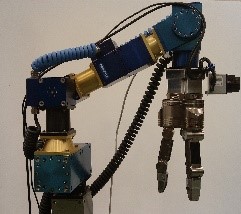} & \includegraphics[width=.28\linewidth,height=60pt]{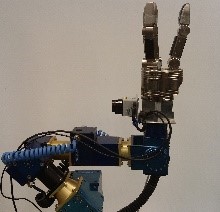} & \includegraphics[width=.28\linewidth,height=60pt]{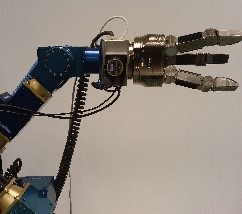}\\
 a) vertically down & b) vertically up & c) horizontal    
 \end{tabular}
\caption{Using a robotic arm to orient SDH in different directions.}
\centering
\label{arm_holding_hand_in_different_angles}
\end{figure}

The hand is mounted on a robotic arm, so we can control its orientation. Three different orientations (vertically up, vertically down and horizontal) were used in the experiments as shown in  Fig ~\ref{arm_holding_hand_in_different_angles}. This orienting allows us to include the effect of gravity as well. It also mimics the versatile hand orientations in object transfer tasks.

\begin{figure}[h]
\centering
\includegraphics[width=0.5\textwidth]{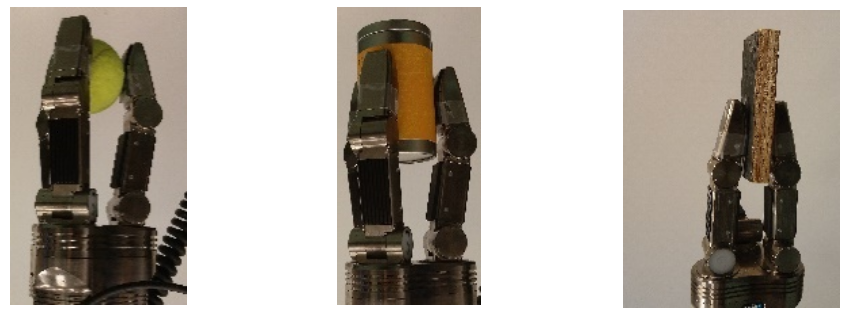}
\centering
\caption{Different objects used in experiment.}
\label{different_objects_used_in_experiment}
\end{figure}

Three different objects were used in the experiment: a tennis ball, a cylindrical cardboard tube and a cuboid plank of wood, as shown in Fig~\ref{different_objects_used_in_experiment}. The contact surfaces for the objects are respectively fiber, paper and unpolished wood. These different object shapes were chosen to span different modalities of contact. For instance, the contact points on a sensor pad  shift as the ball moves or slips away, but they stay relatively stationary for the plank of wood and the cylinder. The plank of wood usually has a biased contact pressure during push or grip, which is due to the fact that a  human subject would usually apply a small torque on the plank during the hand-over task. Using different objects makes event identification more challenging, yet more realistic, since it introduces  variability similar to that found in real applications. Through preliminary experiments, we realised that, for instance, an optical-flow-based method might be robust enough to detect slippage when using a ball object, but it would not serve as a robust method for identifying slippage in other object types.

\begin{figure}[h]
\centering
\includegraphics[width=0.25\textwidth]{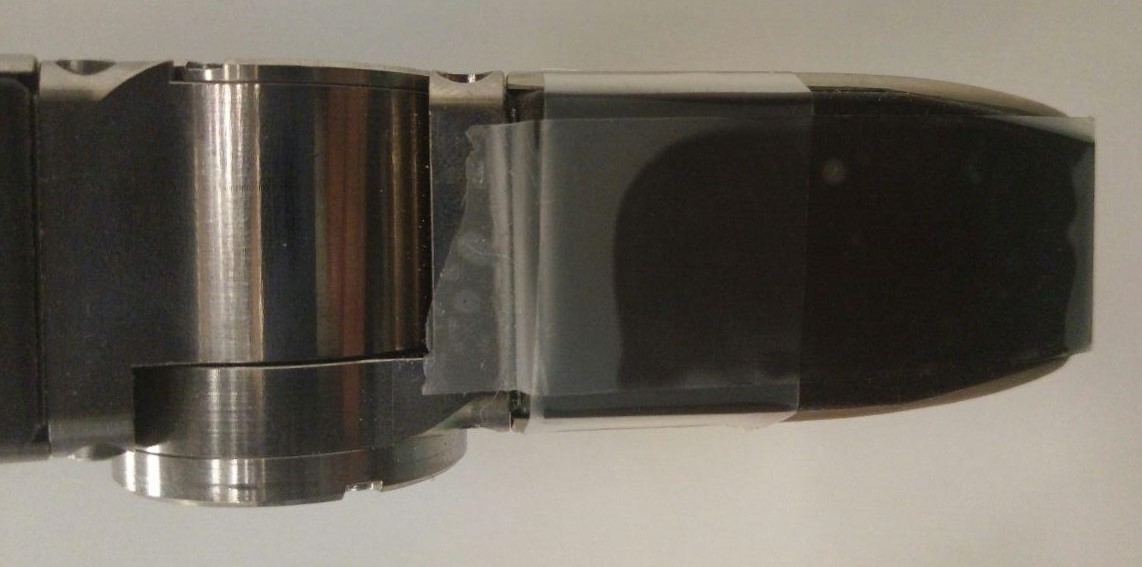}
\caption{Protecting the tactile sensor with tape.}
\centering
\label{protecting_the_hand_with_tape}
\end{figure}

The tactile sensors were covered with tape (Fig~\ref{protecting_the_hand_with_tape}) to perform experiments without damaging the hand or wearing it. The tape also makes slipping easier and ensures a smooth movement of the object on the surface of the finger by decreasing friction, thus creating consistent data stream. We have reasons to believe that our algorithm would perform more accurately and at a higher confidence level if the tapes were removed.

In each experiment, the data was acquired for 3 seconds. A subset of length T (less than 1 second)  was used for training and testing data for each event. Choosing the T value depends on the trade-off between classification performance and the classification speed. For real time application of our approach, the lower the $T$ value, the better it is because less acquisition time is needed for classification. The start of the subset is determined when at least one of the taxel values for all pads exceeds a threshold, set manually in advance, at about 15\% of the maximum taxel value. We used a relatively low threshold to  avoid discarding samples falsely. Manual thresholding can be avoided by introducing a fifth event corresponding to ``dormant'' in the classification scheme.

\section{EXPERIMENTAL PROCEDURE }\label{sec:4}

In order to perform the experiment, 5 graduate students, 4 males and 1 female, between 20-30 years old, volunteered as subjects. The purpose of the test was explained to the students beforehand. They were also shown example videos of each experiment so that they understand the logistics of the experiment clearly. Finally, the subjects were asked to perform each of the following actions  while the object is held in a precision grasp by the SDH:

\begin{enumerate}
\item pull: the subject pulls the object from the robotic hand. Subjects were free to either leave the object in the robot's hand or take it out.
\item push: the subject pushes the object into the robotic hand. Subjects were asked to stop pushing the object when the object loses contact with  the robot’s distal pads.
\item hold (also called grip): the subject grips and holds the object. They were instructed that they could grip in a way that felt natural for them.
\item bump: the subject taps any side of the object once with the distal part of their fingers (to avoid inadvertent damage to the SDH).
\end{enumerate}

\begin{figure}[thpb]
\centering
\includegraphics[width=0.5\textwidth]{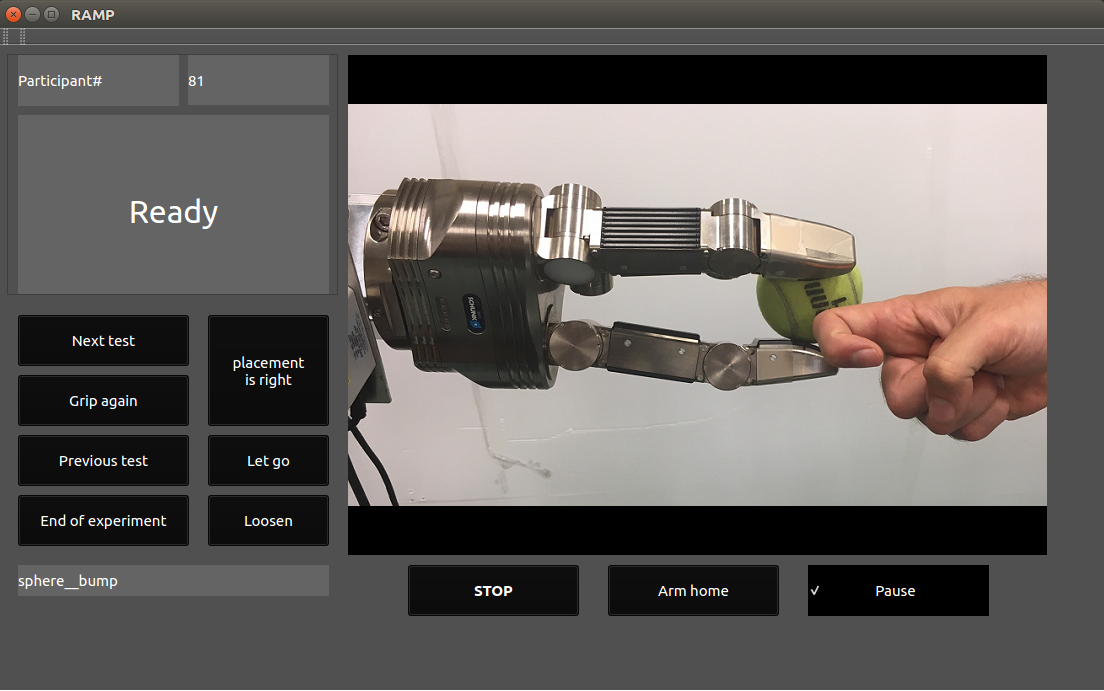}\centering
\caption{The GUI developed for conducting experiments and data acquisition. This figure shows a sample bump test for demonstration to the subjects.}
 \centering
     \label{software}
\end{figure}

A specific GUI, as shown in Fig~\ref{software}, was designed  for data acquisition. The software automates hand opening and closing and the movement of the arm; informs the subject when to start interacting with the object and when to replace the object using voice commands.  In order to have versatile data for training, the participants were free to choose their own hand configuration to perform the action required. Finally, the software saves the tactile data stream for each person in a separate file.

The sequence of experiments is as follows. At the start, the investigator initializes the software and the arm moves to a preset pose and SDH  opens up to receive the ball object.  A pre-recorded video pops up on the monitor showing a sample of a person pushing the object. The investigator  places the ball in the SDH and it closes, thereby holding the ball in a firm grasp. The software signals for the subject to push the object. After 3 seconds, the software signals the subject to stop pushing. After about a second, the software  signals the subject to re-position the ball in the SDH and then signals the subject to push the ball again into the hand. This is repeated 5 times. The same sequence is then applied to pull, hold and bump actions. Next, the entire sequence is repeated for the cylinder and plank objects. Next the robot arm moves to two other poses, and the entire procedure repeats for each pose. Thus, overall we collected 5 persons $\times$ 5 runs per person $\times$ 3 hand poses $\times$ 3 objects $\times$ 4 actions = 900 samples.

\section{RESULTS} \label{sec:5} 

BoW algorithm has three key parameters, K, the dimension of the feature space; W, the window length; and T, the stream length. Later, we show the effect of theses parameters on the accuracy of classification, but first, we present results with the values for which we achieved the most accuracy. These were K = 10, W = 7 and T = 15. Both W and T are measured in number of time steps.  Each time-step is .03 seconds which corresponds to the output frequency (32 Hz) of the sensor hardware. In seconds, therefore, T $\approx$ 0.5 seconds.

\begin{figure}[thpb]
\centering
\includegraphics[width=0.5\textwidth]{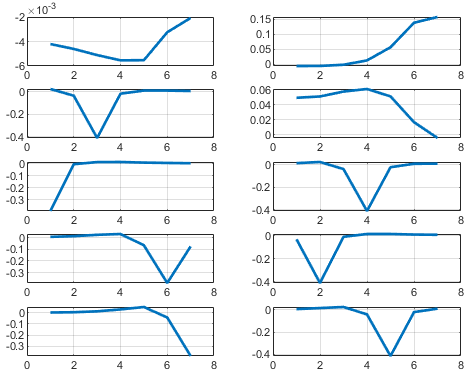}
\centering
\caption{10 centers determined by K-means clustering, each represented as a time series of length W = 7. }
\centering
\label{features}
\end{figure}

\begin{figure}[thpb]
\centering
\includegraphics[width=0.5\textwidth]{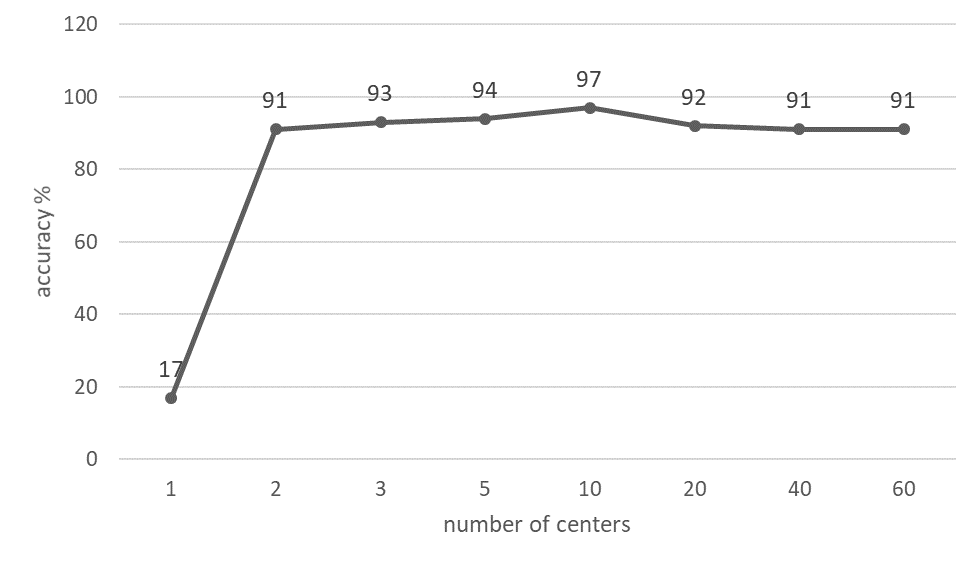}
\centering
\caption{Effect of number of centers, K on accuracy}
      \label{centers}
\end{figure}

\begin{figure}[thpb]
\centering
\includegraphics[width=0.5\textwidth]{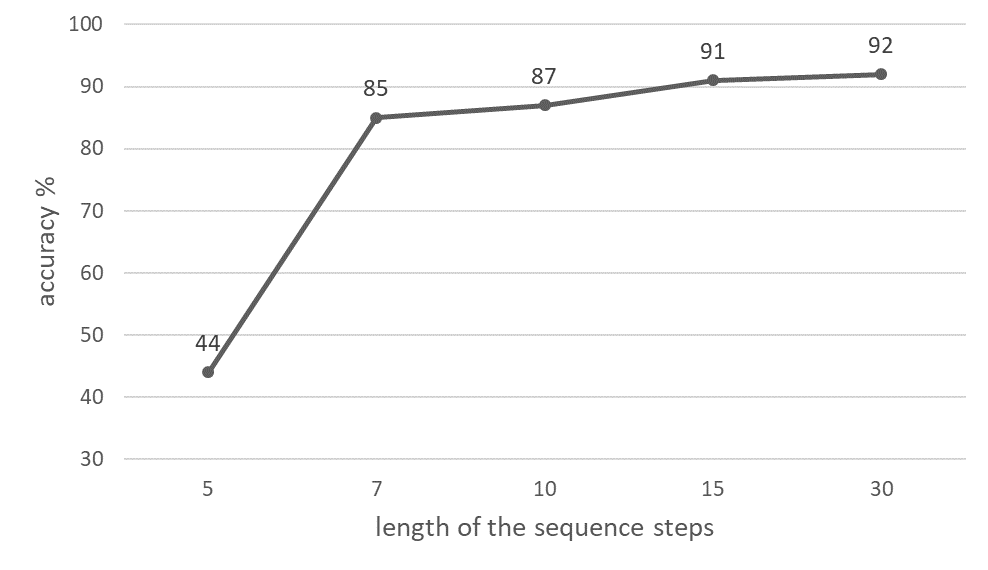}
\centering
\caption{Effect of sequence length, T on accuracy}
      \label{sequence_length}
\end{figure}

\begin{figure}[thpb]
\centering
\includegraphics[width=0.5\textwidth]{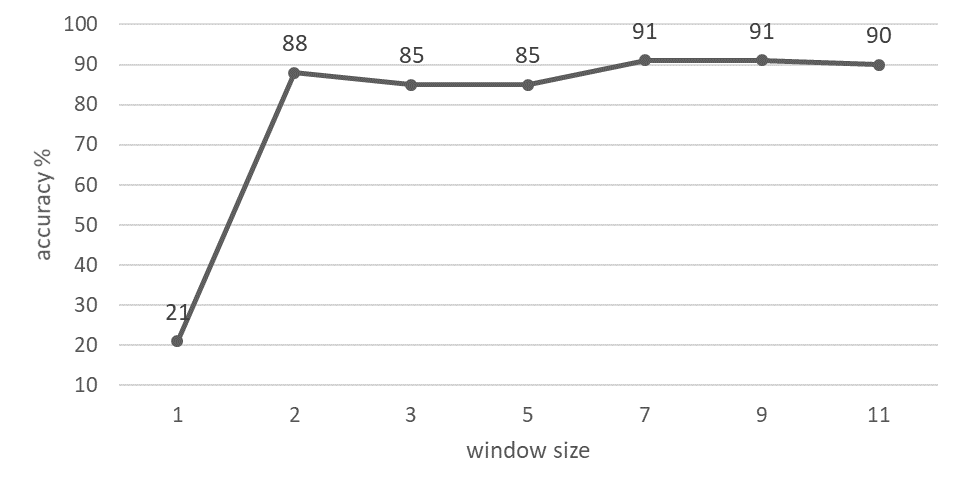}
\centering
\caption{Effect of window size, W on accuracy}
      \label{window_size}
\end{figure}

Fig~\ref{features} shows the different features produced by the BoW algorithm. Average accuracy (vertical axes in Fig \ref{centers}-\ref{window_size}) is calculated by dividing total number of successful classifications in all the classes by the total number of classifications.

Furthermore, the SVM was trained using i) one-vs-one binary classifiers for classification and ii) assuming a hard margin.  The plots shown in this paper are for an SVM trained with a Gaussian kernel. However, we found that a linear SVM also works as well as the one with a Gaussian kernel for the features generated by the BoW algorithm with properly tuned parameters. This is likely due to the fact that the features produced by the BoW result in linearly separable features. 

We arranged the training and test data sets for our BoW+SVM algorithm in two different configurations: 

\begin{enumerate}
\item Data of all 5 subjects is pooled together and the training and test sets are chosen randomly with 80\% of the whole data  used for training and 20\%  for testing (without substitution). The low percentage of test data is due to low number of overall trials. 
\item The training data is chosen from 4 participants and tested on the fifth one, resulting in 5 such combinations. 
\end{enumerate}

\begin{figure}[thpb]
\centering
\includegraphics[width=0.5\textwidth]{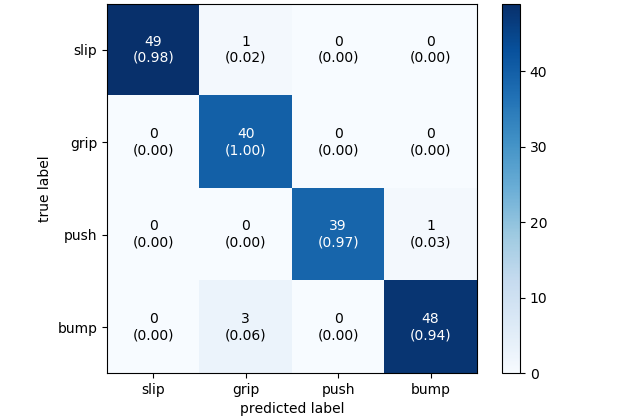}
\centering
\caption{Confusion matrix for Configuration 1 Test where 80\% of data is randomly chosen for training and the rest 20\% of data is used for testing.}
\centering
      \label{confusion_matrix}
\end{figure}

\begin{figure}[thpb]
\centering
\includegraphics[width=0.5\textwidth]{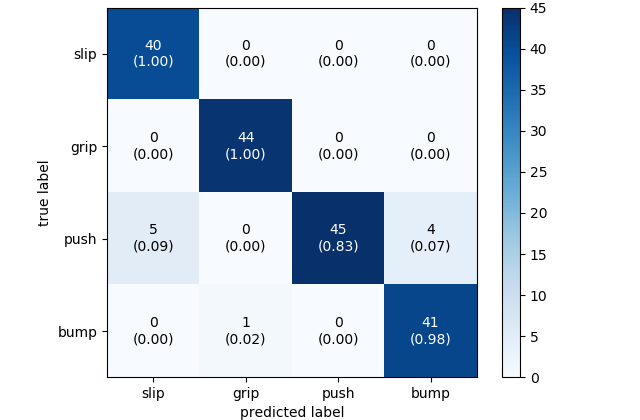}
\centering
\caption{Confusion matrix for Configuration 2 Test where training data is chosen from 4 participants and tested on the fifth one.}
      \label{confusion2}
\end{figure}

Fig~\ref{confusion_matrix} shows the confusion matrix for the first configuration. The results for the second configuration  are shown in Fig~\ref{confusion2}. This shows that the  classifier has learned a rich set of features and can classify events for interaction with a subject who is not in the training set.

As mentioned earlier, the three key parameters in our BoW algorithm are: K, W and T. We now report the effects of varying these parameters.

 K is the number of words or clusters outputted by the k-means sub-algorithm embedded within the main algorithm. Using too many clusters results in over fitting and using too few clusters results in reduced discrimination capability. Fig~\ref{centers} shows the effect of changing K  on accuracy of classification, while keeping W and T constant at 7 and 15, respectively.  K = 10 yields the best results and this explains our use of K = 10 in our BoW algorithm.

For real time event classification in real object transfer scenarios, data stream length T  should not be too long. However, too short a sequence may not give rich enough features. Fig~\ref{sequence_length} shows the effect of T (measured in number of time-steps) on accuracy of the classification while keeping K and W constant (10 and 7, respectively). This explains our use of T = 15 time steps in our experiments which, as mentioned before, corresponds to a duration of about 0.5 seconds for the data stream. 

The sliding window size, W is also important for creating effective features. Fig~\ref{window_size} shows the effect of changing the window size on accuracy of classification, while keeping K and T constant at 10 and 15, respectively. This explains our use of W = 7 in our BoW algorithm.

\begin{figure}[thpb]
\centering
\includegraphics[width=0.5\textwidth]{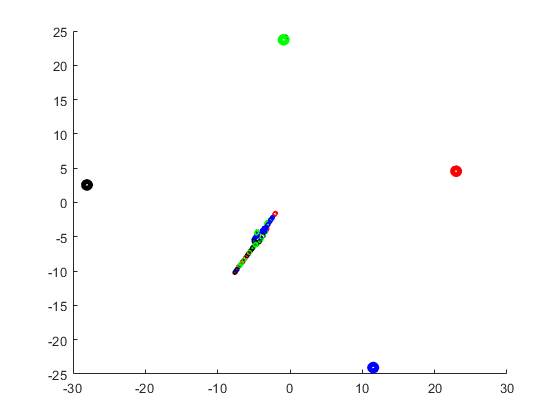}
\centering
\caption{t-SNE representation of the features generated from different experiments. It shows separability of 4 groups of data:  slip (red), grip (black), bump (blue) and push (green).}
\centering
\label{tsne}
\centering
\end{figure}

Finally, to visualize the separability of features, we show the t-SNE \cite{maaten2008visualizing} representation of the features generated by the BoW algorithm in Fig~\ref{tsne}. The t-SNE is essentially an unsupervised algorithm that projects high dimensional data to lower dimensions (two in our case here) in such a way that similar data remain clustered together and dissimilar data is separated. It  is commonly used for visualization purposes for high dimensional data. Each class is assigned a unique colour label and the data points in the figure are colored corresponding to their respective classes.  It is evident that the BoW features are making it likely for the SVM to separate most of the data. There are 4 clearly separated clusters, one corresponding to each class, and one cluster formed from different classes. Please note that this mixed group is larger in this visualization  because the t-SNE algorithm is unsupervised and has no information about the class tags. A supervised algorithm such as the SVM, will generally have a significantly lower misclassification.

\section{CONCLUSIONS and Future Work}\label{sec:6}

In this paper we showed that using only tactile data, it is possible to classify if an object held in a robot's hand is being pushed, pulled, held, or bumped by a human receiver during a robot-human object transfer task. These events are closely related and can commonly occur during an object transfer task. Our core algorithm uses standard machine learning techniques - a BoW algorithm for automated feature generation on the temporal tactile data stream, followed by an SVM for classification. We also empirically determined the best values for the key parameters in the BoW algorithm that result in  about  95\rpm 2\% average classification accuracy.

The experiments we reported in this paper were from a preliminary study. Our next step is to carry out a bigger and more formal study with a larger number of participants and more objects with different geometries and texture to confirm our preliminary findings. Subsequently we intend to apply this method to a robot-human object transfer task, by combining it with  an autonomous fetch and delivery system with a mobile manipulator \cite{mikesposter} being developed in RAMP Lab at SFU.

\bibliography{IEEEexample}

\bibliographystyle{IEEEtran}


\end{document}